\documentclass{article}

\usepackage[T1]{fontenc}

\usepackage{tgtermes}

\usepackage{amsmath}
\usepackage{amsfonts}

\usepackage{scalefnt,letltxmacro}
\usepackage{bm}
\LetLtxMacro{\oldtextsc}{\textsc}
\renewcommand{\textsc}[1]{\oldtextsc{\scalefont{1.10}#1}}
\usepackage[scaled=0.92]{PTSans}
\usepackage{inconsolata}

\usepackage[usenames,dvipsnames]{xcolor}
\definecolor{shadecolor}{gray}{0.9}

\usepackage[final, expansion=alltext]{microtype}
\usepackage[english]{babel}
\usepackage{afterpage}
\usepackage{framed}
\usepackage{nicefrac}

\DeclareRobustCommand{\parhead}[1]{\textbf{#1}~}

\usepackage{graphicx}
\usepackage[labelfont=bf]{caption}
\usepackage[format=hang]{subcaption}

\usepackage{booktabs}

\usepackage{natbib}

\usepackage[algoruled]{algorithm2e}
\usepackage{listings}
\usepackage{fancyvrb}
\fvset{fontsize=\normalsize}

\usepackage[colorlinks, linktoc=all]{hyperref}
\usepackage[all]{hypcap}
\hypersetup{citecolor=Violet}
\hypersetup{linkcolor=black}
\hypersetup{urlcolor=MidnightBlue}

\usepackage[acronym,smallcaps,nowarn]{glossaries}

\newcommand{\red}[1]{\textcolor{BrickRed}{#1}}

\newcommand{\green}[1]{\textcolor{OliveGreen}{#1}}
\newcommand{\blue}[1]{\textcolor{MidnightBlue}{#1}}
\newcommand{\purple}[1]{\textcolor{Plum}{#1}}

\usepackage{listings}
\lstdefinestyle{mystyle}{
    commentstyle=\color{OliveGreen},
    numberstyle=\tiny\color{black!60},
    stringstyle=\color{BrickRed},
    basicstyle=\ttfamily\scriptsize,
    breakatwhitespace=false,
    breaklines=true,
    captionpos=b,
    keepspaces=true,
    numbers=none,
    numbersep=5pt,
    showspaces=false,
    showstringspaces=false,
    showtabs=false,
    tabsize=2
}
\usepackage[normalem]{ulem}

 \DeclareRobustCommand{\mb}[1]{\ensuremath{\boldsymbol{\mathbf{#1}}}}

\newcommand{\cll}{\ensuremath{\mathcal{L}_{\text{pos}}}}

\newcommand{\mbx}{\mb{x}}

\newcommand{\bbR}{\mathbb{R}}

\newacronym{KL}{kl}{Kullback-Leibler}
\newacronym{ELBO}{elbo}{evidence lower bound}
\newacronym{SVI}{svi}{stochastic variational inference}
\newacronym{GMM}{gmm}{Gaussian mixture model}
\newacronym{LDA}{lda}{latent Dirichlet allocation}
\newacronym{PCA}{pca}{principal component analysis}

\newacronym{sgd}{sgd}{stochastic gradient descent}
\newacronym{semb}{s-emb}{static embeddings}
\newacronym{temb}{t-emb}{time-binned embeddings}
\newacronym{demb}{d-emb}{dynamic embeddings}
\newacronym{cbow}{cbow}{continuous bag-of-words}

\newacronym{efe}{efe}{exponential family embedding}
\newacronym{pmi}{pmi}{pointwise mutual information}
 
\usepackage{blindtext}

\usepackage[accepted]{icml2017}

\begin{document}

\twocolumn[
\icmltitle{Dynamic Bernoulli Embeddings for Language Evolution}
\begin{icmlauthorlist}
\icmlauthor{Maja Rudolph,}{}
\icmlauthor{David Blei}{}
\\
\vskip 0.1in
\icmlauthor{Columbia University, New York, USA}{}
\end{icmlauthorlist}
\vskip 0.3in
]

\begin{abstract}
  Word embeddings are a powerful approach for unsupervised analysis of
  language. Recently, \citet{rudolph2016exponential} developed
  exponential family embeddings, which cast word embeddings in a
  probabilistic framework. Here, we develop \textit{dynamic
    embeddings}, building on exponential family embeddings to capture
  how the meanings of words change over time.  We use dynamic
  embeddings to analyze three large collections of historical texts:
  the U.S. Senate speeches from 1858 to 2009, the history of computer
  science ACM abstracts from 1951 to 2014, and machine learning papers
  on the Arxiv from 2007 to 2015.  We find dynamic embeddings provide
  better fits than classical embeddings and capture interesting
  patterns about how language changes.
\end{abstract}

\section{Introduction}

Word embeddings are a collection of unsupervised learning methods for
capturing latent semantic structure in language.  Embedding methods
analyze text data, learning distributed representations of the
vocabulary to capture its co-occurrence statistics.  These learned
representations are then useful for reasoning about word usage and
meaning \citep{harris1954distributional, rumelhart1986learning}.  With
large data sets and approaches from neural networks, word embeddings
have become an important tool for analyzing language
\citep{bengio2003neural, mikolov2013linguistic,
  mikolov2013distributed, mikolov2013efficient, pennington2014glove,
  levy2014neural, arora2015rand}.

Recently, \citet{rudolph2016exponential} developed \textit{exponential
  family embeddings}.  Exponential family embeddings distill the key
assumptions of an embedding problem, generalize them to many types of
data, and cast the distributed representations as latent variables in
a probabilistic model.  They encompass many existing methods for
embeddings and open the door to bringing expressive probabilistic
modeling~\citep{bishop2006machine,murphy2012machine} to the problem of
learning distributed representations \citep{bengio2003neural}.

Here we use exponential family embeddings to develop \textit{dynamic
  word embeddings}, a method for learning distributed representations
that change over time.  Dynamic embeddings analyze long-running texts,
e.g., documents that span many years, where the way words are used
changes over time.  The goal of dynamic embeddings is to characterize
and understand those changes.

\begin{figure}[t]
  \centering
  \hspace*{-2pt}
  \begin{tabular}{c}
  \includegraphics[width=0.43\textwidth]{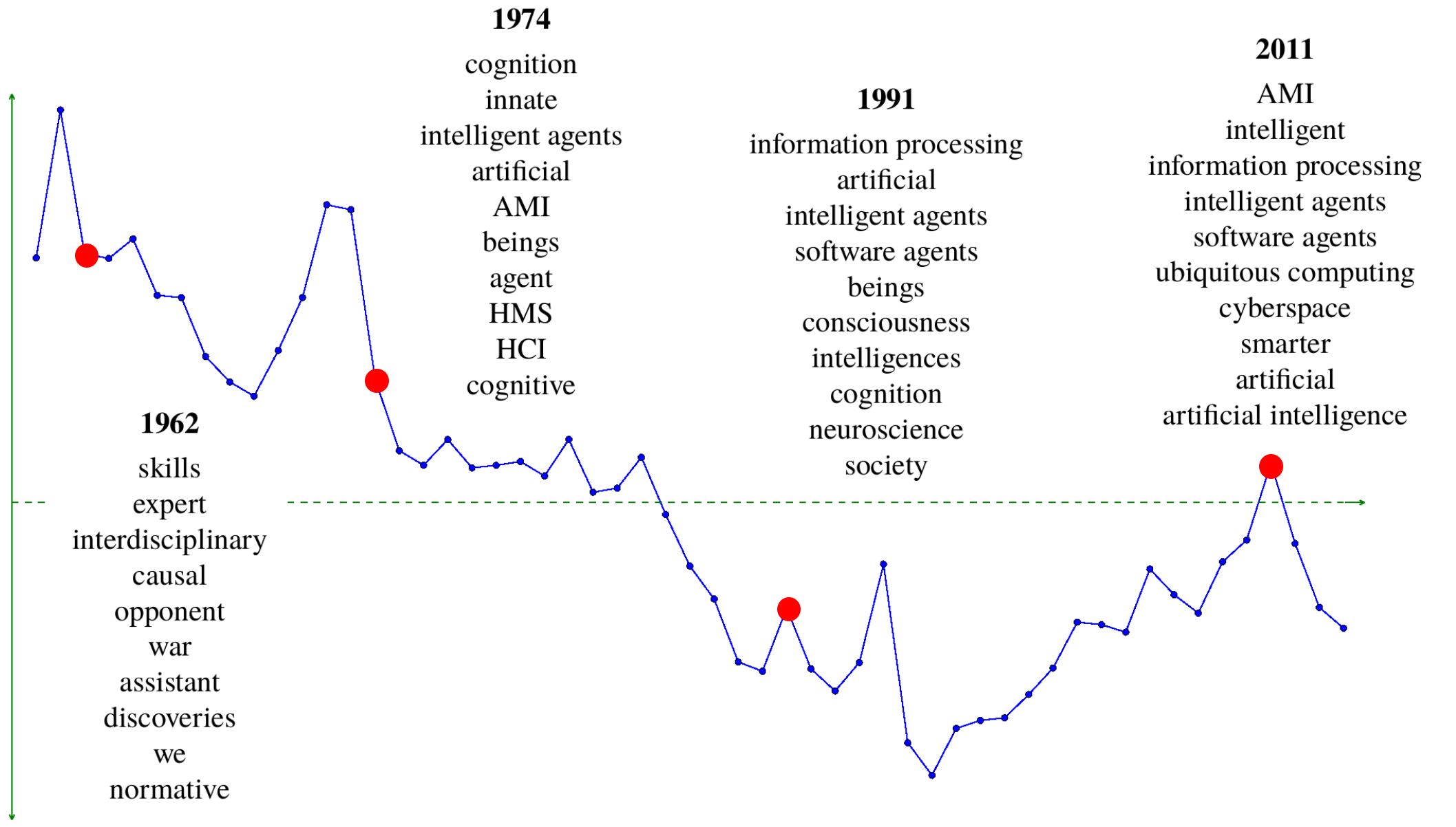} \\
    (a) \textsc{intelligence} in ACM abstracts (1951--2014) \\
    \\
    \includegraphics[width=0.47\textwidth]{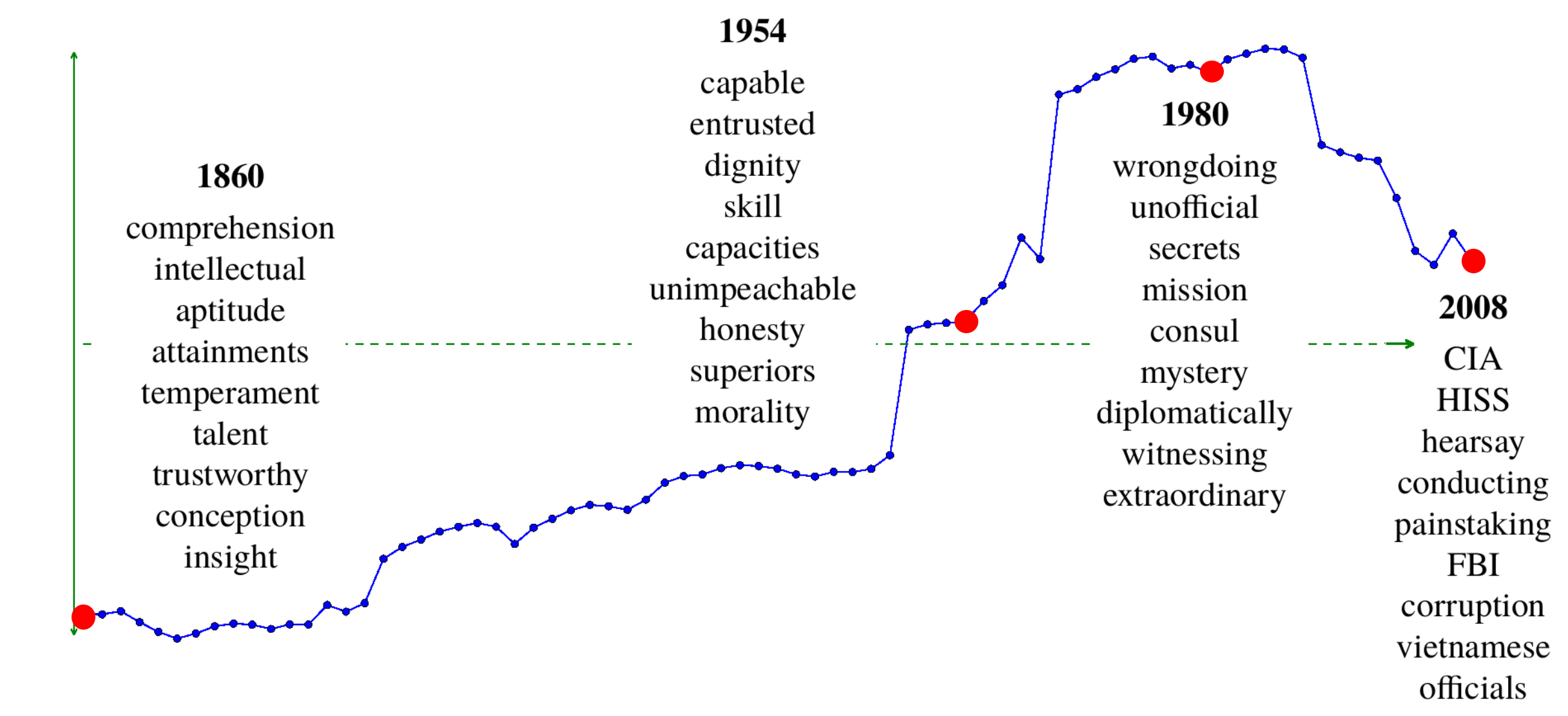} \\
    (b)  \textsc{intelligence} in U.S. Senate speeches (1858--2009) \\
  \end{tabular}
  \caption{The dynamic embedding of \textsc{intelligence} reveals how
    the term's usage changes over the years. The $y$-axis is
    ``meaning,'' a one dimensional projection of the embedding
    vectors. For selected years, we list words with similar dynamic
    embeddings.}
  \label{fig:intelligence}
\end{figure}

Figure\nobreakspace \ref {fig:intelligence} illustrates the approach.  It shows the
changing representation of \textsc{intelligence} in two corpora, the
collection of computer science abstracts from the ACM 1951--2014 and
the U.S. Senate speeches 1858--2009.  On the y-axis is ``meaning,'' a
proxy for the dynamic representation of the word; in both corpora, its
representation changes dramatically over the years.  To understand
where it is located, the plots also show similar words (according to
their changing representations) at various points.  Loosely, in the
ACM corpus \textsc{intelligence} changes from government intelligence
to cognitive intelligence to artificial intelligence; in the
Congressional record \textsc{intelligence} changes from psychological
intelligence to government intelligence.  Section\nobreakspace \ref {sec:empirical} gives
other examples from these corpora, such as \textsc{iraq},
\textsc{data}, and \textsc{computer}.

In more detail, a word embedding uses representation vectors to
parameterize the conditional probabilities of words in the context of
other words.  Dynamic embeddings divide the documents into time
slices, e.g., one per year, and cast the embedding vector as a latent
variable that drifts via a Gaussian random walk.  When fit to data,
the dynamic embeddings capture how the representation of each word
drifts from slice to slice.

Section\nobreakspace \ref {sec:model} describes dynamic embeddings and how to fit them.
Section\nobreakspace \ref {sec:empirical} studies this approach on three datasets: 9 years
of Arxiv machine learning papers (2007--2015), 64 years of computer
science abstracts (1951--2014), and 151 years of U.S. Senate speeches
(1858--2009).  Dynamic embeddings give better predictive performance
than existing approaches and provide an interesting exploratory window
into how language changes.

\parhead{Related work.} Language is known to evolve
\citep{aitchison2001language,kirby2007innateness} and there have been several lines of research around capturing semantic shifts.
\citet{mihalcea2012word,tang2016semantic} detect semantic changes of words using features such as part-of-speech tags and entropy and
\citet{sagi2011tracing, basile2014analysing} employ latent semantic analysis and temporal semantic indexing for quantifying changes in meaning.

Most closely related to our work are methods for dynamic word embeddings
\citep{kim2014temporal,kulkarni2015statistically,
  hamilton2016diachronic}. 
These methods train a separate embedding
model for each time slice of the data.  While interesting, this
requires enough data in each time slice such that a high quality
embedding can be trained for each. Further, because each time slice is
trained independently, the dimensions of the embeddings are not
comparable across time; they must use initialization \citep{kim2014temporal} or ad-hoc alignment techniques \citep{kulkarni2015statistically,hamilton2016diachronic,zhang2016past} to stitch them
together.

In contrast, the representations of dynamic embeddings are sequential
latent variables.  Dynamic embeddings naturally accommodates time
slices with sparse data and immediately connect the latent dimensions
across time.  In Section\nobreakspace \ref {sec:empirical}, we found that dynamic
embeddings provide quantitative improvements over independently
fitting each slice.\footnote{Two similar models have been
  independently developed.
  \citet{bamler2017dynamic} model both the embeddings and the context
  vectors using an Uhlenbeck-Ornstein process
  \citep{uhlenbeck1930theory}. \citet{yao2017discovery} factorize the
  \gls{pmi} matrix at different time slices.
  Their regularization also resembles an Uhlenbeck-Ornstein process.}

Dynamic topic modeling also studies text data over time
\citep{blei2006dynamic,wang2006topics, Wang:2008,
  Gerrish:2010,wijaya2011understanding,Yogatama:2014,mitra2014s,mitra2015automatic,frermann2016bayesian}.
This class of models describes documents in terms of topics, which are
distributions over the vocabulary, and then allows the topics to
change over the course of the collection.  As in dynamic embeddings,
some dynamic topic models use a Gaussian random walk to capture drift
in the underlying language model; for example, see
\citet{blei2006dynamic,Wang:2008,Gerrish:2010,frermann2016bayesian}.

Though topic models and word embeddings are related, they are
ultimately different approaches to language modeling.  Topic models
capture co-occurrence of words at the document level and focus on
heterogeneity, i.e., that a document can exhibit multiple topics
\citep{blei2003latent}.  Word embeddings capture co-occurrence in
terms of proximity in the text, usually focusing on small
neighborhoods around each word \citep{mikolov2013linguistic}.
Combining dynamic topic models and dynamic word embeddings is an area
for future study.

 \section{Dynamic Embeddings}
\label{sec:model}

We develop dynamic embeddings, a type of \gls{efe}
\citep{rudolph2016exponential} that captures sequential changes in the
representation of the data.  We focus on text data and the Bernoulli
embedding model.

In this section, we review Bernoulli embeddings for text and show how
to include dynamics into the model.  We then derive the objective
function for dynamic embeddings and develop stochastic gradients to
optimize it.

\parhead{Bernoulli embeddings for text.} An exponential family
embedding is a conditionally specified model.  It has three
ingredients: The {\em context}, the {\em conditional distribution} of
each data point, and the {\em parameter sharing structure}.

In an \gls{efe} for text, the data is a corpus of text. It is a
collection of words $(x_1, \ldots, x_N)$ from a vocabulary of size
$V$.  Each word $x_i \in \{0,1\}^V$ is an indicator vector (also
called a ``one-hot'' vector).  It has exactly one nonzero entry at
$v$, where $v$ is the vocabulary term at position $i$.

In an \gls{efe} each data point has a \textit{context}. In text, the
context of each word is its neighborhood; thus it is modelled
conditional on the words that come before and after it.  Typical
context sizes range between $2$ and $10$ words. (This is set in
advance or by cross-validation.)

We will build on Bernoulli embeddings, which provide a conditional
model for the individual entries of the indicator vectors
$x_{iv}\in\{0,1\}$. Let $c_{i}$ be the set of positions in the
neighborhood of position $i$ and let $\mbx_{c_i}$ denote the
collection of data points indexed by those positions.  The conditional
distribution of $x_{iv}$ is
\begin{align}
  \label{eqn:bernoulli}
  x_{iv} | \mbx_{c_{i}} \sim \text{Bern}(p_{iv}),
\end{align}
where $p_{iv} \in (0,1)$ is the Bernoulli
probability.\footnote{Multinomial embeddings
  \citep{rudolph2016exponential} model each indicator vector $x_i$
  with a categorical conditional distribution, but this requires
  expensive normalization in form of a softmax function.  For
  computational efficiency, one can replace the softmax with the
  hierarchical softmax \citep{mikolov2013distributed} or employ
  approaches related to noise contrastive estimation
  \citep{gutmann2010noise,mnih2013learning}.  Bernoulli embeddings
  relax the one-hot constraint of $x_i$, and work well in practice;
  they relate to the negative sampling
  \citep{mikolov2013distributed}.}

Bernoulli embeddings specify the natural parameter, which is the log
odds $\eta_{iv} = \log \frac{p_{iv}}{1-p_{iv}}$.  It is a function of
the representation of term $v$ and the terms in the context of
position $i$.  Specifically, each index $(i,v)$ in the data is
associated with two parameter vectors, the {\it embedding vector}
$\rho_v \in \bbR^K$ and the {\it context vector}
$\alpha_v \in \bbR^K$.  Together, the embedding vectors and context
vectors form the natural parameter of the Bernoulli.  It is
\begin{align}
  \label{eqn:logitparams}
  \eta_{iv} = \rho_v^\top \left(\textstyle \sum_{j \in c_{i}} \sum_{v'} \alpha_{v'} x_{jv'}\right).
\end{align}
This is the inner product between the embedding $\rho_v$ and the
context vectors of the words that surround position $i$.  (Because
$x_{j}$ is an indicator vector, the sum over the vocabulary selects
the appropriate context vector $\alpha$ at position $j$.)  The goal is
to learn the embeddings and context vectors.

The index on the parameters does not depend on position $i$, but only
on term $v$; the embeddings are shared across all positions in the
text.  This is what \citet{rudolph2016exponential} call the
\textit{parameter sharing structure}.  It ensures, for example, that
the embedding vector for \textsc{intelligence} is the same wherever it
appears in the corpus.  (Dynamic embeddings partially relax this
restriction.)

Finally, \citet{rudolph2016exponential} regularize the Bernoulli
embedding by placing priors on the embedding and context vectors.
They use Gaussian priors with diagonal covariance, i.e., $\ell_2$
regularization.  Without the regularization, fitting a Bernoulli
embedding closely relates to other embedding techniques such as CBOW
\citep{mikolov2013efficient} and negative sampling
\citep{mikolov2013distributed}.  But the probabilistic perspective of
\citet{rudolph2016exponential}---and in particular the priors and the
parameter sharing---allows us to extend this setting to capture
dynamics.

\begin{figure}[h!]
\centering
\includegraphics[width=0.5\textwidth]{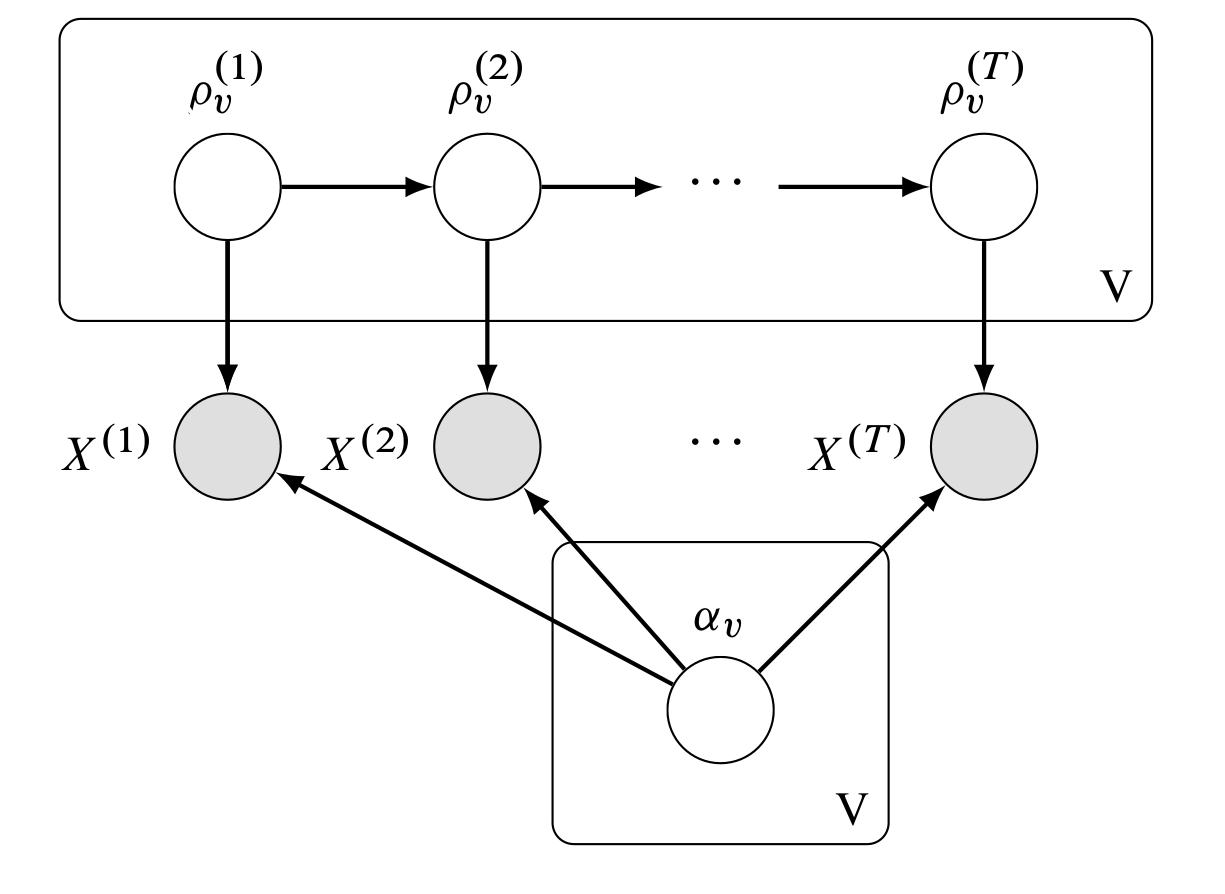}
\caption{Graphical representation of a dynamic embedding for text data in $T$ time slices, $X^{(1)},\cdots, X^{(T)}$. The embedding vectors $\rho_v$ of each term evolve over time. The context vectors are shared across all time slices.}
\label{fig:graphical_model}
\end{figure}
 
\parhead{Dynamic Bernoulli embeddings.}  Dynamic Bernoulli embeddings
extend Bernoulli embeddings to text data over time. Each observation
$x_{iv}$ is associated with a time slice $t_i$, such as the year of
the observation.  Context vectors are shared across all positions in
the text but the embedding vectors are only shared within a time
slice.  Thus dynamic embeddings posit a sequence of embeddings for
each term $\rho^{(t)}_v\in\bbR^{K}$.

The natural parameter of the conditional likelihood is similar to
Equation\nobreakspace \textup {(\ref {eqn:logitparams})} but with the embedding vector $\rho_v$ replaced
by the per-time-slice embedding vector $\rho_v^{(t_i)}$,
\begin{align}
  \label{eqn:logitsdyn}
  \eta_{iv} = \rho_v^{(t_i)\top} \left(\textstyle \sum_{j \in c_{j}} \sum_{v'} \alpha_{v'} x_{jv'}\right).
\end{align}

Finally, dynamic embeddings use a Gaussian random walk as a prior on
the embedding vectors,
\begin{align}
  \label{eqn:prior}
  \alpha_v, \rho_v^{(0)} &\sim \mathcal{N} (0, \lambda_{0}^{-1} I) \\
  \rho_v^{(t)} &\sim \mathcal{N} (\rho_v^{(t-1)}, \lambda^{-1} I).
\end{align}
Given data, this leads to smoothly changing estimates of each term's
embedding.\footnote{Because $\bm{\alpha}$ and $\bm{\rho}$ appear only
  as inner products in ~Equation\nobreakspace \textup {(\ref {eqn:logitparams})}, there is some
  redundancy in placing temporal dynamics on both the embeddings and
  the context vectors. Exploring dynamics in $\bm{\alpha}$ is a
  subject for future study.}

Figure\nobreakspace \ref {fig:graphical_model} gives the graphical model for dynamic
embeddings.  Dynamic embeddings are a conditionally specified model,
which in general are not guaranteed to imply a consistent joint
distribution.  But dynamic Bernoulli embeddings model binary data, and
thus a joint exists \citep{arnold2001conditionally}.

\parhead{Fitting dynamic embeddings.} Calculating the joint is
computationally intractable.  Rather, we fit dynamic embeddings with
the \textit{pseudo log likelihood}, the sum of the log conditionals.
This is a commonly used objective for conditionally specified models
\citep{arnold2001conditionally}.

In detail, we regularize the pseudo log likelihood with the log priors
and then maximize to obtain a pseudo MAP estimate. For dynamic
Bernoulli embeddings, this objective is the sum of the log priors and
the conditional log likelihoods of the data $x_{iv}$.  We divide the
data likelihood into two parts, the contribution of nonzero data
entries $\mathcal{L}_{\text{pos}}$ and contribution of zero data
entries $\mathcal{L}_{\text{neg}}$,
\begin{align}
  \label{eqn:objective}
  \mathcal{L}(\bm{\rho}, \bm{\alpha}) &= \mathcal{L}_{\text{pos}} +
                                        \mathcal{L}_{\text{neg}} +
                                        \mathcal{L}_{\text{prior}}.
\end{align}

The likelihoods are
\begin{align*}
  \mathcal{L}_{\text{pos}} &=  \sum_{i=1}^N \sum_{v=1}^Vx_{iv}\log \sigma(\eta_{iv}) \nonumber \\
  \mathcal{L}_{\text{neg}} &=      \sum_{i=1}^N \sum_{v=1}^V( 1 - x_{iv}) \log ( 1 - \sigma(\eta_{iv})), \nonumber \\
\end{align*}
where $\sigma(\cdot)$ is the sigmoid, which maps natural parameters to
probabilities.

The prior is
\begin{align*}
  \mathcal{L}_{\text{prior}} &= \log p(\bm{\alpha}) + \log
                               p(\bm{\rho}),
\end{align*}
where
\begin{align*}
  \log p(\bm{\alpha})
  &=-\frac{\lambda_{0}}{2}
    \sum_v
    || \alpha_v||^2 \\
  \log p(\bm{\rho})
  &= -\frac{\lambda_{0}}{2}
    \sum_v
    ||\rho_v^{(0)}||^2
    -\frac{\lambda}{2}
    \sum_{v,t}
    ||\rho_v^{(t)}-\rho_v^{(t-1)}||^2.
\end{align*}

The parameters $\bm{\rho}$ and $\bm{\alpha}$ appear in the natural
parameters $\eta_{iv}$ of Equations\nobreakspace \textup {(\ref {eqn:logitparams})} and\nobreakspace  \textup {(\ref {eqn:logitsdyn})} and in
the log prior.  The random walk prior penalizes consecutive word
vectors $\rho_v^{(t-1)}$ and $\rho_v^{(t)}$ for drifting too far
apart.  It prioritizes parameter settings for which the norm of their
difference is small.

The most expensive term in the objective is
$\mathcal{L}_{\text{neg}}$, the contribution of the zeroes to the
conditional log likelihood.  The objective is cheaper if we subsample
the zeros.  Rather than summing over all words which are not at
position $i$, we sum over a subset of negative examples drawn at
random.  \citet{mikolov2013distributed} call this negative sampling
and recommend sampling from the unigram distribution raised to the
power of $0.75$.

With negative sampling, we redefine $\mathcal{L}_{\text{neg}}$ in
Equation\nobreakspace \textup {(\ref {eqn:objective})}.  Denote the sampling distribution of zeros as
$\hat{p}$,
\begin{align}
  \mathcal{L}_{\text{neg}} =  \sum_{i=1}^N \sum_{v\sim \hat{p}}\log ( 1 - \sigma(\eta_{iv})).
\label{eqn:negativesampling}
\end{align}
This sum has fewer terms and reduces the contribution of the zeros to
the objective.  In a sense, this incurs a bias---the expectation with
respect to the negative samples is not equal to the original
objective---but ``downweighting the zeros'' can improve prediction
accuracy \citep{hu2008collaborative,liang2016modeling}.

We fit the objective (Equation\nobreakspace \textup {(\ref {eqn:objective})} with
Equation\nobreakspace \textup {(\ref {eqn:negativesampling})}) using stochastic gradients
\citep{robbins1951stochastic} and with adaptive learning rates
\citep{duchi2011adaptive}. Pseudo code is in Appendix\nobreakspace \ref {sec:pseudo}. To avoid deriving the gradients of
Equation\nobreakspace \textup {(\ref {eqn:objective})}, we implemented the algorithm in Edward
\citep{tran2016edward}.  Edward is based on tensorflow
\citep{tensorflow2015whitepaper} and employs automatic
differentiation.\footnote{Code available at \url{http://github.com/mariru/dynamic\_bernoulli\_embeddings}}

 \begin{table}[]
\centering
\scriptsize
\caption{Time range and size of the three corpora analyzed in Section\nobreakspace \ref {sec:empirical}.}
\label{tab:data}
\begin{tabular}{lccccc}
\\
\toprule
         & {\bf time range}  & {\bf slices} & {\bf slice size} & {\bf vocab size} & {\bf words} \\
{\bf Arxiv ML} & $2007 - 2015$ & $9$ & 1 year  & $50$k     & $6.5$M \\
{\bf ACM} & $1951 - 2014$ & $64$ &1 year & $25$k     & $21.6$M \\
{\bf Senate speeches} & $1858 - 2009$ & $76$ &2 years & $25$k     & $13.7$M\\
\bottomrule
\end{tabular}
\end{table}
\section{Empirical Study}
\label{sec:empirical}
Our empirical study contains two parts. In a quantitative evaluation
we benchmark dynamic embeddings against static embeddings
\citep{mikolov2013efficient,mikolov2013distributed,rudolph2016exponential}.
Dynamic embeddings improve over static embeddings in terms of the
conditional likelihood of held-out predictions.  Further, dynamic
embeddings perform better than embeddings trained on the individual
time slices \citep{hamilton2016diachronic}.  In a qualitative
evaluation we use a fitted dynamic embedding model to extract which
word vectors change most and we visualize their dynamics.  Dynamic
embeddings provide a new window into how language changes.

\subsection{Data}

We studied three datasets. See Table\nobreakspace \ref {tab:data}.

{\it Machine Learning Papers (2007 - 2015):} This dataset contains the
full text from all machine learning papers (tagged ``stat.ML'')
published on the Arxiv between April 2007 and June 2015.  It spans 9
years and we treat each year as a time slice. The number of Arxiv
papers about machine learning has increased over the years.  There
were $101$ papers in $2007$; there were $1,573$ papers in $2014$.

{\it Computer Science Abstracts (1951 - 2014):} This dataset
contains abstracts of computer science papers published by the
Association of Computing Machinery (ACM) from 1951 to 2014.  Again,
each year is considered a time slice and here too the amount of data
increases over the years.  For 1953, there are only around $10$ abstracts and their combined length is only $471$ words; the combined length of the abstracts from 2009 is over 2M.

{\it Senate Speeches (1858 - 2009):} This dataset contains all
U.S. Senate speeches from 1858 to mid 2009.  Here we treat every 2
years as a time slice. In contrast to the other datasets, this corpus
is a transcript of spoken language.

For all datasets, we divide the observations into training,
validation, and testing. Within each time slice we use $80 \%$ for
training, $10 \%$ for validation, and $10 \%$ for testing.
Appendix\nobreakspace \ref {sec:preprocessing} provides details about preprocessing.

\subsection{Quantitative evaluation}

We compare \gls{demb} to \gls{temb} \citep{hamilton2016diachronic} and
\gls{semb} \citep{rudolph2016exponential}.  There are many embedding
techniques, without dynamics, that enjoy comparable performance.  For
the \gls{semb}, we study Bernoulli embeddings
\citep{rudolph2016exponential}, which are similar to \gls{cbow} with
negative sampling \citep{mikolov2013efficient,mikolov2013distributed}.
For time-binned embeddings, \citet{hamilton2016diachronic} train a
separate embedding on each time slice.

\parhead{Evaluation metric.} We evaluate models by held-out Bernoulli
probability.  Given a model, each held-out word (validation or
testing) is associated with a Bernoulli probability.  At that
position, a better model assigns higher probability to the observed
word and lower probability to the others.  This metric is
straightforward because the competing methods all produce Bernoulli
conditional likelihoods (Equation\nobreakspace \textup {(\ref {eqn:bernoulli})}).\footnote{Since we hold
  out chunks of consecutive words usually both a word and its context
  are held out. For all methods we have to use the words in the
  context to compute the conditional likelihoods.}  We report \cll,
which considers only the nonzero held-out data.  To make results
comparable, all methods are trained with the same number of negative
samples.

\parhead{Model training and hyperparameters.}  Each method takes a
maximum of $10$ passes over the data.  (The corresponding number of
stochastic gradient steps depends on the size of the minibatches.)
The parameters of \gls{semb} are initialized randomly.  We initialize
both \gls{demb} and \gls{temb} from a fit of \gls{semb} which has been
trained from one pass, and then train for $9$ additional passes.

We set the dimension of the embeddings to $100$ and the number of
negative samples to $20$.  We experiment with two context sizes, $2$
and $8$.

Other parameters are set by validation error.  All methods use
validation error to set the initial learning rate $\eta$ and minibatch
sizes $m$.  The model selects $\eta \in [0.01,0.1,1,10]$ and
$m \in [0.001 N, 0.0001 N,0.00001 N]$, where $N$ is the size of
training data.  The only parameter specific to \gls{demb} is the
precision of the random drift.  To have one less hyper parameter to tune, 
we fix the precision on the context vectors and the initial dynamic embeddings to
$\lambda_{0} = \lambda/1000$, a constant multiple of the precision on the dynamic embeddings.
We choose $\lambda \in [1, 10]$ by validation error.

\parhead{Results.}  We train each model on each training set and use
each validation set for model selection (e.g., selecting the minibatch
size and learning rate).  Table\nobreakspace \ref {tab:results} reports the results on
the test set.  Dynamic embeddings consistently achieve higher held-out
likelihood.

\begin{table}[h!]
\scriptsize
\centering
\caption{  \glsreset{semb}
  \glsreset{temb}
  \glsreset{demb}
  \Gls{demb} consistently achieve highest held-out
  $\mathcal{L}_{\text{pos}}$ (Equation\nobreakspace \textup {(\ref {eqn:objective})}). We compare to
  \gls{semb} \citep{mikolov2013distributed,rudolph2016exponential},
  \gls{temb} \citep{hamilton2016diachronic}.}
\label{tab:results}
\begin{tabular}{lcc}
\\
\multicolumn{3}{c}
{\bf Arxiv ML}  
\\ \toprule
                         & context size $2$      & context size $8$      \\
\gls{semb} \citep{rudolph2016exponential}         &   $-2.706 \pm 0.002$     &  $-2.491 \pm 0.002$      \\
\gls{temb} \citep{hamilton2016diachronic}       &   $-2.646\pm 0.002$     &  $-2.454 \pm 0.002$      \\
\gls{demb} [this paper] &   $\bm{-2.535 \pm 0.001}$     & $\bm{-2.400 \pm 0.002 }$    \\
\bottomrule
\\
\multicolumn{3}{c}
{\bf Senate speeches}   
\\ \toprule
                         & context size $2$      & context size $8$      \\
\gls{semb}  \citep{rudolph2016exponential}        &   $-2.366 \pm 0.001$     &  $-2.244 \pm 0.001$      \\
\gls{temb}  \citep{hamilton2016diachronic}      &   $-2.295 \pm 0.001$     &  $-2.212 \pm 0.001$      \\
\gls{demb} [this paper] &   $\bm{-2.263 \pm 0.001}$     & $\bm{-2.204 \pm 0.001}$    \\
\bottomrule
\\
\multicolumn{3}{c}
{\bf ACM}  
\\ \toprule
                         & context size $2$      & context size $8$      \\
\gls{semb} \citep{rudolph2016exponential}      &   $-2.427 \pm 0.001$     &  $-2.231 \pm 0.001$      \\
\gls{temb}  \citep{hamilton2016diachronic}   &   $-2.420 \pm 0.001$     &  $-2.242 \pm 0.001$      \\
\gls{demb} [this paper] &   $\bm{-2.396 \pm 0.001}$     &  $\bm{-2.228 \pm 0.001}$      \\
\bottomrule
\end{tabular}
\end{table}

\subsection{Qualitative exploration}

We now show how to use dynamic embeddings to explore the dataset. We
use the fitted model to suggest ways that language changes and
visualize its discovered dynamic structure.

A word's {\it embedding
  neighborhood} helps visualize its usage and how it changes over
time. It is simply a list of other words with similar usage. For a
given query word (e.g., \textsc{computer}) we take its index $v$ and
select the top ten words according to
\begin{align}
  \label{eqn:nbr} \text{neighborhood}(v, t) = \text{argsort}_{w}\big(
  \frac{\text{sign}(\rho_v^{(t)})^{\top}\rho^{(t)}_w}{||
  \rho^{(t)}_v||\cdot||\rho^{(t)}_w||}\big).
\end{align}

\begin{table}[t!]
\scriptsize
\centering
\caption{Embedding neighborhoods (Equation\nobreakspace \textup {(\ref {eqn:nbr})}) reveal how the usage of a word changes over time. The embedding neighborhoods of \textsc{computer} and \textsc{bush} were computed from a dynamic embedding fitted to Congress speeches (1858-2009). \textsc{computer} used to be a profession but today it is used to refer to the electonic device. The word \textsc{bush} is a plant but eventually in congress \textsc{Bush} is used to refer to the political figures. The embedding neighborhood of \textsc{data} comes from a dynamic embedding fitted to ACM abstracts (1951-2014).}
\label{tab:example_drift}
\begin{minipage}{.2\textwidth}
\begin{tabular}{cc}
\\
\multicolumn{2}{c}
{\bf \textsc{computer} (Senate)}\\
\toprule
{\bf 1858 }       &{\bf  1986}        \\
 computer        &   computer              \\
 draftsman      &    software             \\
 draftsmen      &    computers            \\
 copyist        &    copyright            \\
 photographer   &    technological        \\
 computers      &    innovation           \\
 copyists       &    mechanical           \\
 janitor         &   hardware              \\
 accountant      &   technologies          \\
 bookkeeper      &   vehicles              \\
\bottomrule
\end{tabular}
\end{minipage}\hspace{25pt} 
\begin{minipage}{.2\textwidth}
\begin{tabular}{cc}
\\
\multicolumn{2}{c}
{\bf \textsc{bush} (Senate)}\\
\toprule
{\bf 1858 }       &{\bf  1990}        \\
 bush            &   bush                  \\
 barberry       &    cheney               \\
 rust           &    nonsense             \\
 bushes         &    nixon                \\
 borer          &    reagan               \\
 eradication    &    george               \\
 grasshoppers   &    headed               \\
 cancer          &   criticized            \\
 tick            &   clinton               \\
 eradicate       &   blindness             \\
\bottomrule
\end{tabular}
\end{minipage}\hfill
\begin{minipage}{.5\textwidth}
\begin{tabular}{ccccc}
\\
\multicolumn{5}{c}
{\bf \textsc{data} (ACM)}\\
\toprule
{\bf 1961 }       & {\bf 1969}              & {\bf 1991}       &  {\bf 2011}     & {\bf 2014}      \\
data              &  data                   &  data            &  data           &  data                       \\          
directories       &  repositories           &  voluminous      &  raw data       &  data streams               \\
files             &  voluminous             &  raw data        &  voluminous     &  voluminous                 \\
bibliographic     &  lineage                &  repositories    &  data sources   &  raw data                   \\
formatted         &  metadata               &  data streams    &  data streams   &  warehouses                 \\
retrieval         &  snapshots              &  data sources    &  dws            &  dws                        \\
publishing        &  data streams           &  volumes         &  repositories   &  repositories               \\
archival          &  raw data               &  dws             &  warehouses     &  data sources               \\
archives          &  cleansing              &  dsms            &  marts          &  data mining                \\
manuscripts       &  data mining            &  data access     &  volumes        &  marts                      \\
\bottomrule
\end{tabular}
\end{minipage}\hfill
\end{table}
 
\begin{figure}[]
  \centering
  \hspace*{-10pt}
  \includegraphics[width=0.45\textwidth]{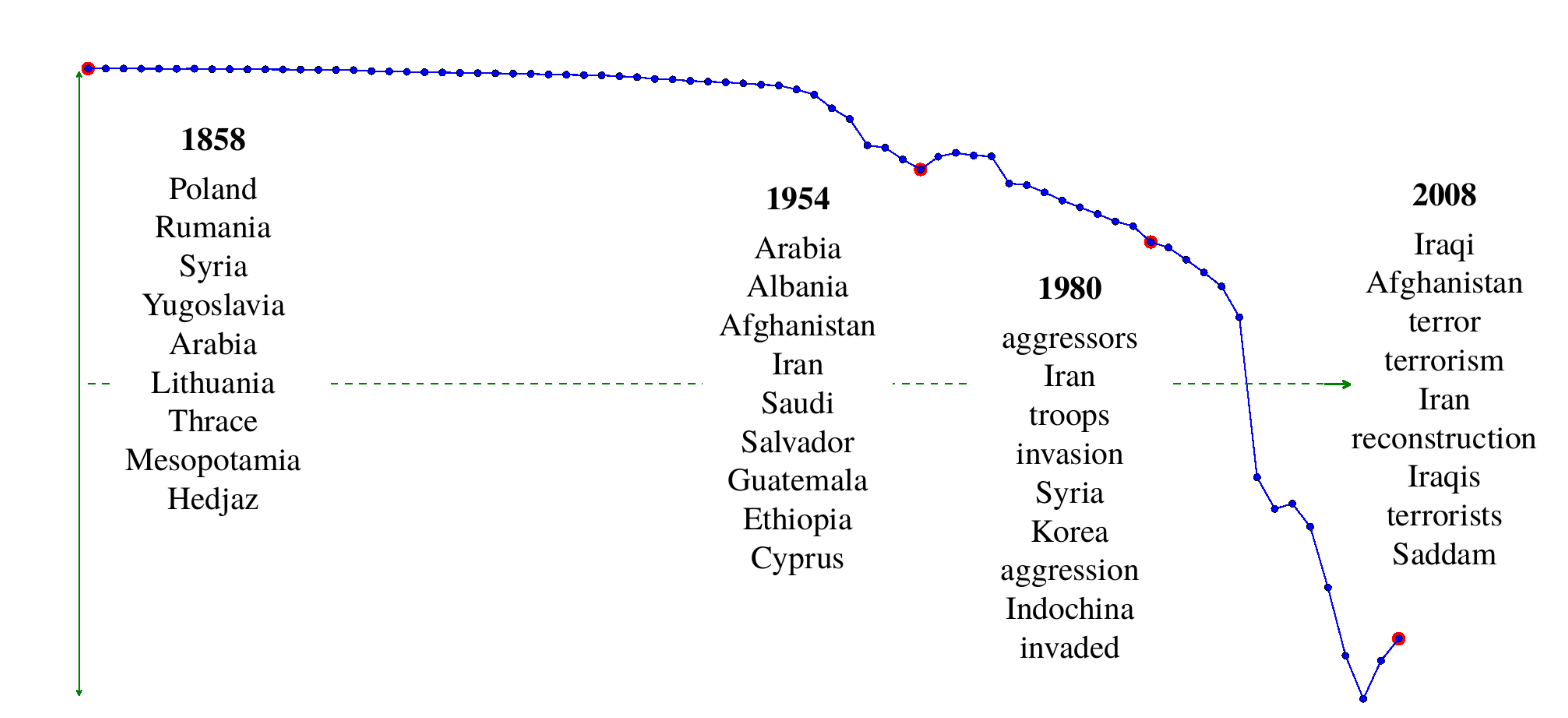}
  \caption{The dynamic embedding captures how the usage of the word
    \textsc{iraq} changes over the years (1858-2009). The $x$-axis is
    time and the $y$-axis is a one-dimensional projection of the
    embeddings using \gls{PCA}. We include the embedding neighborhoods
    for \textsc{Iraq} in the years 1858, 1954, 1980 and 2008.}
\label{fig:iraq}
\end{figure}

We fit a dynamic embedding fit to the Senate speeches.
Table\nobreakspace \ref {tab:example_drift} gives the embedding neighborhoods of
\textsc{computer} for the years $1858$ and $1986$.  Its usage changed
dramatically over the years. In 1858, a \textsc{computer} was a
profession, a person who was hired to compute things.  Now the
profession is obsolete; \textsc{computer} refers to the electronic
device.

Table\nobreakspace \ref {tab:example_drift} provides another example, \textsc{bush}.  In
1858 this word always referred to the plant.  A \textsc{bush} still is
a plant, but in the 1990's, in the Senate, it is usually refers to
political figures.  Unlike \textsc{computer}, where the embedding
neighborhoods reveal two mutually exclusive meanings, the embedding
neighborhoods of \textsc{bush} reflect which meaning is more prevalent
in a given period.

A final example in Table\nobreakspace \ref {tab:example_drift} is the word \textsc{data},
from the ACM abstracts.  The evolution of the embedding neighborhoods
of \textsc{data} reflects how its meaning changes in the computer
science literature.

\parhead{Finding changing words with absolute drift.}  We have
highlighted example words whose usage changes.  However, not all words
have changing usage.  We now define a metric to discover which words
change most.

\begin{table}[]
\scriptsize
\centering
\caption{A list of the top 16 words whose dynamic embedding on Senate
  speeches changes most. The number represents the absolute drift
  (Equation\nobreakspace \textup {(\ref {eqn:drift})}).
  The dynamics of the capitalized words are in
  Table\nobreakspace \ref {tab:topdrift_topics} and discussed in the main text. }
\label{tab:congress_topdrift}
\begin{tabular}{llll}
\\
\multicolumn{4}{c}
{\bf words with largest drift (Senate)}\\
\toprule
\textsc{iraq }        &  3.09 &   coin               &  2.39  \\
tax cuts           &  2.84 &   social security    &  2.38  \\
health care        &  2.62 &   \textsc{fine}         &  2.38  \\
energy             &  2.55 &   signal             &  2.38  \\
medicare           &  2.55 &   program            &  2.36  \\
\textsc{discipline}   &  2.44 &   moves              &  2.35  \\
text               &  2.41 &   credit             &  2.34  \\
\textsc{values}       &  2.40 &   \textsc{unemployment} &  2.34  \\
\bottomrule
\end{tabular}
\end{table}
\begin{table*}[]
\scriptsize
\centering
\caption{Embedding neighborhoods extracted from a dynamic embedding fitted to Senate speeches (1858 - 2009). \textsc{discipline}, \textsc{values}, \textsc{fine}, and \textsc{unemployment} are within the 16 words whose dynamic embedding has largest absolute drift. (Table\nobreakspace \ref {tab:congress_topdrift}).}
\label{tab:topdrift_topics}
\begin{minipage}{.16\textwidth}
\begin{tabular}{cc|}
\\
\multicolumn{2}{c}
{\bf \textsc{discipline}}\\
\toprule
{\bf 1858 }       &{\bf  2004}        \\
 discipline      &  discipline              \\
 hazing         &   balanced               \\
 westpoint      &   balancing              \\
 assaulting     &   fiscal                 \\
 disciplined    &   let                    \\
 courtmartial   &   ourselves              \\
 punishment     &   structural             \\
 martial         &  deficit                 \\
 mentally        &  administrations         \\
 summarily       &  restraint               \\
\bottomrule
\end{tabular}
\end{minipage}\hfill
\begin{minipage}{.18\textwidth}
\begin{tabular}{cc|}
\\
\multicolumn{2}{c}
{\bf \textsc{values}}\\
\toprule
{\bf 1858 }       &{\bf  2000}        \\
 values               &    values                    \\
 fluctuations        &     sacred                   \\
 value               &     inalienable              \\
 currencies          &     unique                   \\
 fluctuation         &     preserving               \\
 depreciation        &     exemplified              \\
 fluctuating         &     principles               \\
 purchasing power     &    philanthropy              \\
 fluctuate            &    virtues                   \\
 basis                &    historical                \\
\bottomrule
\end{tabular}
\end{minipage}\hfill
\begin{minipage}{.16\textwidth}
\begin{tabular}{cc|}
\\
\multicolumn{2}{c}
{\bf \textsc{fine}}\\
\toprule
{\bf 1858 }       &{\bf  2004}        \\
 fine              &  fine              \\
 luxurious        &   punished         \\
 finest           &   penitentiaries   \\
 coarse           &   imprisonment     \\
 beautiful        &   misdemeanor      \\
 imprisonment     &   punishable       \\
 finer            &   offense          \\
 lighter           &  guilty            \\
 weaves            &  conviction        \\
 spun              &  penitentiary      \\
\bottomrule
\end{tabular}
\end{minipage}\hfill
\begin{minipage}{.33\textwidth}
\begin{tabular}{ccc}
\\
\multicolumn{3}{c}
{\bf \textsc{unemployment}}\\
\toprule
{\bf 1858 }      &{\bf  1940}  & {\bf 2000}      \\
 unemployment        &  unemployment   &   unemployment    \\
 unemployed          &  unemployed      &  jobless         \\
 depression          &  depression      &  rate            \\
 acute               &  alleviating     &  depression      \\
 deplorable          &  destitution     &  forecasts       \\
 alleviating         &  acute           &  crate           \\
 destitution         &  reemployment    &  upward          \\
 urban               &  deplorable     &   lag             \\
 employment          &  employment     &   economists      \\
 distressing         &  distress       &   predict         \\
\bottomrule
\end{tabular}
\end{minipage}
\end{table*}
 \begin{table*}[]
\scriptsize
\centering
\caption{Using dynamic embeddings we can study a social phenomenon of interest. We pick a target word of interest, such as \textsc{jobs} or \textsc{prostitution} and create their embedding neighborhoods (Equation\nobreakspace \textup {(\ref {eqn:nbr})}). Looking at the neighborhood of \textsc{jobs} complements the evolution of \textsc{unemployment} (Table\nobreakspace \ref {tab:topdrift_topics}). Or we might want to study \textsc{prostitution}. It used to be considered immoral and vile, evolved to be indecent and its neighborhood in 1990 reveals a more concerned outlook as it includes words like \textsc{servitude}, \textsc{harassment} and \textsc{trafficking}.}
\label{tab:social_topics}
\hfill
\begin{minipage}{.3\textwidth}
\begin{tabular}{ccc}
\scriptsize
\\
\multicolumn{3}{c}
{\bf \textsc{jobs}}\\
\toprule
{\bf 1858 }       & {\bf 1938}     &  {\bf  2008}        \\
 jobs             & jobs           &  jobs               \\
 employment       & unemployed     &  job                \\
 unemployed       & employment     &  create             \\
 overtime         & job            &  creating           \\
 positions        & overtime       &  tremendously       \\
 job              & positions      &  economies          \\
 idleness         & shifts         &  opportunities      \\
 working          & idleness       &  created            \\
 busy             & busy           &  pace               \\
 civil service    & salaried       &  michigan           \\
\bottomrule
\end{tabular}
\end{minipage}\hfill
\begin{minipage}{.6\textwidth}
\begin{tabular}{cccccc}
\scriptsize
\\
\multicolumn{6}{c}
{\bf \textsc{prostitution}}\\
\toprule
{\bf 1858 }       & {\bf 1930}     &  {\bf  1945}      & {\bf 1962 }       & {\bf 1988}     &  {\bf  1990}        \\
 prostitution    & prostitution      &  prostitution   &  prostitution       & harassment        &  prostitution         \\
 punishing      &  punishing        &   indecent       &  indecent          &  intimidation     &   servitude           \\
 immoral        &  immoral          &   vile           &  harassment        &  prostitution     &   harassment          \\
 illegitimate   &  bootlegging      &   immoral        &  intimidation      &  counterfeit      &   intimidation        \\
 riotous        &  riotous          &   induces        &  sexual            &  illegal          &   trafficking         \\
 mobs           &  forbidden        &   incite         &  vile              &  trafficking      &   harassing           \\
 violence       &  anarchists       &   abortion       &  counterfeit       &  indecent         &   apprehended         \\
 assemblage      & assemblage        &  forbid         &  anarchists         & disregard         &  killings             \\
 criminals       & forbid            &  harboring      &  mobs               & anarchists        &  labeled              \\
 procures        & abet              &  assemblage     &  lawbreakers        & punishing         &  naked                \\
\bottomrule
\end{tabular}
\end{minipage}
\end{table*}
 
One way to find words that change is to use \textit{absolute drift}.
For word $v$, it is
\begin{align}
  \label{eqn:drift}
  \text{drift}(v) = ||\rho_{v}^{(T)} - \rho_{v}^{(0)}||.
\end{align}
This is the Euclidean distance between the word's embedding at the
last time slice and at the first time slice.

In the Senate speeches, Table\nobreakspace \ref {tab:congress_topdrift} shows the 16
words that have largest absolute drift. The word \textsc{iraq} has
largest drift. Figure\nobreakspace \ref {fig:iraq} highlights \textsc{iraq}'s embedding
neighborhood in four time slices: 1858, 1950, 1980, and 2008. 
(Appendix\nobreakspace \ref {sec:iraq} gives the entire trajectory of its
 embedding neighborhood.)
At first
the neighborhood contains other countries and regions. Later, Arab
countries move to the top of the neighborhood, suggesting that the
speeches start to use rhetoric more specific to Arab countries. In
1980, Iraq invades Iran and the Iran-Iraq war begins. In these years
words such as \textsc{aggressors}, \textsc{troops}, and
\textsc{invasion} appear in the embedding neighborhood. Eventually, by
2008, the neighborhood contains \textsc{terror}, \textsc{terrorism},
and \textsc{saddam}.

Four other words with large drift are \textsc{discipline},
\textsc{values}, \textsc{fine} and \textsc{unemployment}
(Table\nobreakspace \ref {tab:congress_topdrift}). Table\nobreakspace \ref {tab:topdrift_topics} shows their
embedding neighborhoods for selected years.  Of these words,
\textsc{discipline}, \textsc{values} and, \textsc{fine} have multiple
meanings. Their neighborhoods reflect how the dominant meaning changes
over time. For example, \textsc{values} can be either a numerical
quantity or can be used to refer to moral values and principles.  In
contrast, \textsc{iraq} and \textsc{unemployment} are both words which
have always had the same definition. Yet, the evolution of their
neighborhood captures changes in the way they are used.

\parhead{Dynamic embeddings as a tool to study a text.}  Our hope is
that dynamic embeddings provide a suggestive tool for understanding
change in language.  For example, researchers interested in
\textsc{unemployment} can complement their investigation by looking at
the embedding neighborhood of related words such as
\textsc{employment}, \textsc{jobs} or \textsc{labor}. In
Table\nobreakspace \ref {tab:social_topics} we list the neighborhoods of \textsc{jobs}
for the years 1858, 1938, and 2008. In 2008 the embedding neighborhood
contains words like \textsc{create} and \textsc{opportunities},
suggesting a different outlook on \textsc{jobs} than in earlier
years.

Another interesting example is \textsc{prostitution}. It used to be
\textsc{immoral} and \textsc{vile}, went to \textsc{indecent}, and in
modern days it is considered \textsc{harassment}. We note the word
\textsc{prostitution} is not a frequent word.  On average,
it is used once per time slice and, in two thirds of the time
slices, it is not mentioned at all. Yet, the model is able to learn
about \textsc{prostitution} and the temporal evolution of the embedding
neighborhood reveals how over the years a judgemental stance turns
into concern over a social issue.

 \section{Summary}

We described dynamic embeddings, distributed representations of words
that drift over the course of the collection.  Building on
\citet{rudolph2016exponential}, we formulate word embeddings with
conditional probabilistic models and then incorporate dynamics with a
Gaussian random walk prior.  We fit dynamic embeddings with stochastic
optimization.

We used dynamic embeddings to analyze several datasets: 8 years of
machine learning papers, 63 years of computer science abstracts, and
151 years of speeches in the U.S. Senate.  Dynamic embeddings provided
a better fit than static embeddings and other methods that account for
time.

Finally, we demonstrated how dynamic embeddings can help identify
interesting ways that language changes.  A word's meaning can change
(e.g., \textsc{computer}); its dominant meaning can change (e.g.,
\textsc{values}); or its related subject matter can change (e.g.,
\textsc{iraq}).

 \section*{Acknowledgements} 
We would like to thank Francisco Ruiz and Liping Liu for discussion
and helpful suggestiongs, Elliot Ash and Suresh Naidu for access to
the Congress speeches, and Aaron Plasek and Matthew Jones for access
to the ACM abstracts.

\bibliography{bib}
\bibliographystyle{icml2017}

\onecolumn
\appendix
\clearpage{}\section{Data Preprocessing}
\label{sec:preprocessing}
We fix the vocabulary to the $25000$ most frequent words and remove all words from the documents which are not in the vocabulary. As in \citep{mikolov2013distributed} we additionally remove each word with probability $p = 1 - \sqrt(\frac{10^{-5}}{f_i})$ where $f_i$ is the frequency of the word. This effectively downsamples especially the frequent words and speeds up training. From each time slice $80 \%$  of the words are used for training. A random subsample of $10 \%$ of the words is held out for validation and another $10 \%$ for testing. 

\section{Pseudo code}
\label{sec:pseudo}
\begin{algorithm}[h]
   \caption{Minibatch stochastic gradient descent for dynamic Bernoulli embeddings.}
   \label{alg:sgd}
\begin{algorithmic}
   \STATE {\bfseries Input:} $T$ time slices of text data $X^{(t)}$ of size $m_t$ respectively. Context size $c$, size of embedding $K$, number of negative samples $n$, number of minibatch fractions $m$, initial learning rate $\eta$, precision $\lambda$, vocabulary size $V$, smoothed unigram distribution $\hat p$.
   \FOR{$v=1$ {\bfseries to} $V$}
       \STATE Initialize entries of $\alpha_v$ 
       \STATE (using draws from a normal distribution with zero mean and standard deviation $0.01$).
       \FOR{$v=1$ {\bfseries to} $V$}
           \STATE Initialize entries of $\rho_v^{(t)}$
           \STATE (using draws from a normal distribution with zero mean and standard deviation $0.01$).
       \ENDFOR
   \ENDFOR
   \FOR {number of passes over the data}
       \FOR {number of minibatch fractions $m$}
           \FOR{$t=1$ {\bfseries to} $T$}
               \STATE Sample minibatch of $m_t/m$ consecutive words $\{x_1^{(t)}, \cdots, x_{m_t/m}^{(t)}\}$ from each time slice $X^{(t)}$, and use each word's context to construct
               \begin{align}
                 C_{i}^{(t)} =  \sum_{j \in c_{i}} \sum_{v'=1}^V \alpha_{v'} x_{jv'}.
               \end{align}
               \STATE For each text position in the minibatch, draw a set $\mathcal{S}_{i}^{(t)}$ of $n$ negative samples from $\hat p$ 
           \ENDFOR
           \STATE update the parameters $\bm{\theta} = \{\bm{\alpha}, \bm{\rho}\}$ by ascending the stochastic gradient
           \begin{align}
                \nabla_{\bm{\theta}} \Bigg\{&
                    \sum_{t=1}^T m \sum_{i=1}^{m_t/m}
                        \Big( \sum_{v=1}^V 
                            x_{iv}^{(t)}\log \sigma( \rho_v^{(t)\top}C_{i}^{(t)}) 
                    +\sum_{x_j\in\mathcal{S}_i^{(t)}} \sum_{v=1}^V
                            ( 1 - x_{jv}) \log ( 1 - \sigma( \rho_v^{(t)\top}C_{i}^{(t)}))
                        \Big) \nonumber \\
                    & -\frac{\lambda_{0}}{2}
                       \sum_v
                       || \alpha_v||^2 
                     -\frac{\lambda_{0}}{2}
                       \sum_v
                       ||\rho_v^{(0)}||^2
                       -\frac{\lambda}{2}
                       \sum_{v,t}
                       ||\rho_v^{(t)}-\rho_v^{(t-1)}||^2 \Bigg\} \nonumber
           \end{align}
       \ENDFOR
   \ENDFOR
\STATE Any standard gradient-based learning rate schedule can be used. We use Adagrad \citep{duchi2011adaptive} in our experiments.
\end{algorithmic}
\end{algorithm}

\newpage
\section{Entire Embedding Trajectory of \textsc{iraq}}
\label{sec:iraq}
Here we give the entire trajectory of the embedding neighborhood of \textsc{iraq}.
 Over the years it drifts smoothly.  On average
 \textsc{iraq} is mentioned only $10.6$ times per time slice and in 64
 out of the 76 time slices, \textsc{iraq} is not even mentioned at all.
 For these years, the prior (Equation\nobreakspace \textup {(\ref {eqn:prior})}) ensures that the
 embedding at time $t$ is the average of the embeddings at time $t-1$
 and $t+1$. When the embedding vector does not change between two
 consecutive time slices, the embedding neighborhood might still
 fluctuate. This is because computing the embedding neighborhoods
 (Equation\nobreakspace \textup {(\ref {eqn:nbr})}) involves also the embedding vectors of the other
 words in the vocabulary.
\begin{table*}[h]
\label{tab:iraq}
\scriptsize
\tiny
\centering
\caption{
Embedding neighborhood of \textsc{iraq} extracted from a dynamic embedding fitted to the congress data. 
It is the word whose embedding vector has largest absolute drift.
By listing the neighborhood for all the time bins, we can see how Iraq's embedding vector drifts smoothly.
In \green{1858} \textsc{iraq's} embedding neighborhood contains countries and regions. In \blue{1950} a rethoric more specific to arabic countries crystalizes.
In \purple{1980} Iraq invades Iran and words like \purple{invasion}, \purple{aggressor} and \purple{troops} are in the neighborhood. By \red{2008} the embedding neighborhood of iran contains words like \red{terror}, \red{terrorism} and \red{saddam}.}
\label{tab:iraq_drift}
\vspace{10pt} 
\begin{minipage}{.33\textwidth}
\begin{tabular}{ll}
\tabularnewline
{\bf \green{1858} } &   
\green{poland,     
rumania,    
syria,      
yugoslavia, 
arabia,}   \\&  
\green{lithuania,  
thrace,     
mesopotamia,
hedjaz,     
albania}    
\vspace{1.5pt} \\
{\bf 1860 } &   
poland,          
rumania,         
syria,           
yugoslavia,      
arabia,     \\&     
lithuania,       
thrace,          
mesopotamia,     
hedjaz,          
albania         
\vspace{1.5pt}\\
{\bf 1862 } &   
poland,          
rumania,         
syria,           
yugoslavia,      
arabia,       \\&   
lithuania,       
thrace,          
mesopotamia,     
hedjaz,          
albania         
\vspace{1.5pt}\\
{\bf 1864 } &   
poland,          
rumania,         
syria,           
yugoslavia,      
arabia,         \\& 
lithuania,       
thrace,          
mesopotamia,     
hedjaz,          
albania         
\vspace{1.5pt}\\
{\bf 1866 } &   
poland,          
rumania,         
syria,           
yugoslavia,      
arabia,         \\& 
lithuania,       
thrace,          
mesopotamia,     
hedjaz,          
albania         
\vspace{1.5pt}\\
{\bf 1868 } &   
poland,          
rumania,         
syria,           
yugoslavia,      
arabia,         \\& 
lithuania,       
thrace,          
mesopotamia,     
hedjaz,          
albania         
\vspace{1.5pt}\\
{\bf 1870 } &   
poland,          
rumania,         
syria,           
yugoslavia,      
arabia,         \\& 
lithuania,       
thrace,          
mesopotamia,     
hedjaz,          
albania         
\vspace{1.5pt}\\
{\bf 1872 } &   
poland,          
rumania,         
syria,           
yugoslavia,      
arabia,         \\& 
lithuania,       
thrace,          
mesopotamia,     
hedjaz,          
albania         
\vspace{1.5pt}\\
{\bf 1874 } &   
poland,          
rumania,         
syria,           
yugoslavia,      
arabia,         \\& 
lithuania,       
thrace,          
mesopotamia,     
hedjaz,          
albania         
\vspace{1.5pt}\\
{\bf 1876 } &   
poland,          
rumania,         
syria,           
yugoslavia,      
arabia,         \\& 
lithuania,       
thrace,          
mesopotamia,     
hedjaz,          
albania         
\vspace{1.5pt}\\
{\bf 1878 } &   
poland,          
rumania,         
syria,           
yugoslavia,      
arabia,         \\& 
lithuania,       
thrace,          
mesopotamia,     
hedjaz,          
albania         
\vspace{1.5pt}\\
{\bf 1880 } &   
poland,          
rumania,         
syria,           
yugoslavia,      
arabia,         \\& 
thrace,          
lithuania,       
mesopotamia,     
hedjaz,          
albania         
\vspace{1.5pt}\\
{\bf 1882 } &   
poland,          
rumania,         
yugoslavia,      
syria,           
arabia,         \\& 
thrace,          
lithuania,       
mesopotamia,     
hedjaz,          
albania         
\vspace{1.5pt}\\
{\bf 1884 } &   
rumania,         
poland,          
yugoslavia,      
syria,           
arabia,         \\& 
thrace,          
lithuania,       
mesopotamia,     
hedjaz,          
albania         
\vspace{1.5pt}\\
{\bf 1886 } &   
rumania,         
poland,          
yugoslavia,      
syria,           
arabia,         \\& 
thrace,          
lithuania,       
mesopotamia,     
hedjaz,          
albania         
\vspace{1.5pt}\\
{\bf 1888 } &   
rumania,         
poland,          
yugoslavia,      
syria,           
arabia,         \\& 
thrace,          
lithuania,       
mesopotamia,     
hedjaz,          
albania         
\vspace{1.5pt}\\
{\bf 1890 } &   
rumania,         
poland,          
yugoslavia,      
arabia,          
syria,          \\& 
thrace,          
lithuania,       
mesopotamia,     
hedjaz,          
albania         
\vspace{1.5pt}\\
{\bf 1892 } &   
rumania,         
poland,          
yugoslavia,      
arabia,          
syria,          \\& 
thrace,          
lithuania,       
mesopotamia,     
hedjaz,          
albania         
\vspace{1.5pt}\\
{\bf 1894 } &   
rumania,         
yugoslavia,      
arabia,          
poland,          
syria,          \\& 
thrace,          
lithuania,       
mesopotamia,     
hedjaz,          
albania         
\vspace{1.5pt}\\
{\bf 1896 } &   
rumania,         
yugoslavia,      
arabia,          
syria,           
poland,         \\& 
thrace,          
lithuania,       
mesopotamia,     
hedjaz,          
albania         
\vspace{1.5pt}\\
{\bf 1898 } &   
rumania,         
yugoslavia,      
arabia,          
syria,           
poland,         \\& 
thrace,          
lithuania,       
mesopotamia,     
hedjaz,          
albania         
\vspace{1.5pt}\\
{\bf 1900 } &   
rumania,         
yugoslavia,      
arabia,          
syria,           
poland,         \\& 
thrace,          
mesopotamia,     
lithuania,       
albania,         
hedjaz          
\vspace{1.5pt}\\
{\bf 1902 } &   
rumania,         
yugoslavia,      
arabia,          
syria,           
poland,         \\& 
thrace,          
mesopotamia,     
lithuania,       
albania,         
hedjaz          
\vspace{1.5pt}\\
{\bf 1904 } &   
rumania,         
yugoslavia,      
arabia,          
syria,           
poland,         \\& 
thrace,          
mesopotamia,     
albania,         
lithuania,       
hedjaz          
\vspace{1.5pt}\\
{\bf 1906 } &   
rumania,         
yugoslavia,      
arabia,          
syria,           
poland,         \\& 
thrace,          
mesopotamia,     
albania,         
lithuania,       
hedjaz          
\vspace{1.5pt}\\
 & \\ &
\vspace{1.5pt}
\end{tabular}
\end{minipage}\hfill
\begin{minipage}{.33\textwidth}
\begin{tabular}{ll}
\\
{\bf 1908 } &   
rumania,         
yugoslavia,      
arabia,          
syria,           
poland,         \\& 
thrace,          
mesopotamia,     
albania,         
hedjaz,          
lithuania       
\vspace{1.5pt}\\
{\bf 1910 } &   
syria,           
rumania,         
arabia,          
yugoslavia,      
thrace,         \\& 
mesopotamia,     
czecho,          
albania,         
poland,          
lithuania       
\vspace{1.5pt}\\
{\bf 1912 } &   
syria,           
rumania,         
arabia,          
yugoslavia,      
mesopotamia,    \\& 
thrace,          
czecho,          
albania,         
lithuania,       
poland          
\vspace{1.5pt}\\
{\bf 1914 } &   
syria,           
rumania,         
arabia,          
yugoslavia,      
thrace,         \\& 
mesopotamia,     
albania,         
czecho,          
poland,          
lithuania       
\vspace{1.5pt}\\
{\bf 1916 } &   
syria,           
rumania,         
arabia,          
yugoslavia,      
thrace,         \\& 
mesopotamia,     
albania,         
czecho,          
poland,          
lithuania       
\vspace{1.5pt}\\
{\bf 1918 } &   
rumania,         
syria,           
arabia,          
yugoslavia,      
poland,         \\& 
czecho,          
mesopotamia,     
lithuania,       
thrace,          
serbia          
\vspace{1.5pt}\\
{\bf 1920 } &   
rumania,         
syria,           
arabia,          
yugoslavia,      
poland,         \\& 
czecho,          
mesopotamia,     
thrace,          
lithuania,       
serbia          
\vspace{1.5pt}\\
{\bf 1922 } &   
rumania,         
syria,           
arabia,          
yugoslavia,      
poland,         \\& 
mesopotamia,     
czecho,          
thrace,          
lithuania,       
serbia          
\vspace{1.5pt}\\
{\bf 1924 } &   
rumania,         
syria,           
arabia,          
yugoslavia,      
poland,         \\& 
mesopotamia,     
czecho,          
thrace,          
lithuania,       
serbia          
\vspace{1.5pt}\\
{\bf 1926 } &   
rumania,         
arabia,          
syria,           
mesopotamia,     
yugoslavia,     \\& 
albania,         
thrace,          
salvador,        
persian,         
czecho          
\vspace{1.5pt}\\
{\bf 1928 } &   
arabia,          
rumania,         
syria,           
yugoslavia,      
mesopotamia,    \\& 
albania,         
thrace,          
persian,         
salvador,        
bulgaria        
\vspace{1.5pt}\\
{\bf 1930 } &   
arabia,          
rumania,         
syria,           
yugoslavia,      
mesopotamia,    \\& 
albania,         
thrace,          
persian,         
salvador,        
bulgaria        
\vspace{1.5pt}\\
{\bf 1932 } &   
arabia,          
rumania,         
syria,           
yugoslavia,      
mesopotamia,    \\& 
albania,         
thrace,          
salvador,        
persian,         
bulgaria        
\vspace{1.5pt}\\
{\bf 1934 } &   
arabia,          
rumania,         
syria,           
yugoslavia,      
albania,        \\& 
mesopotamia,     
thrace,          
salvador,        
persian,         
bulgaria        
\vspace{1.5pt}\\
{\bf 1936 } &   
arabia,          
rumania,         
syria,           
yugoslavia,      
albania,        \\& 
mesopotamia,     
thrace,          
salvador,        
persian,         
bulgaria        
\vspace{1.5pt}\\
{\bf 1938 } &   
rumania,         
arabia,          
syria,           
yugoslavia,      
albania,        \\& 
mesopotamia,     
bulgaria,        
salvador,        
alsace,          
lithuania       
\vspace{1.5pt}\\
{\bf 1940 } &   
rumania,         
arabia,          
syria,           
yugoslavia,      
mesopotamia,    \\& 
albania,         
salvador,        
guatemala,       
bulgaria,        
thrace          
\vspace{1.5pt}\\
{\bf 1942 } &   
arabia,          
yugoslavia,      
guatemala,       
albania,         
salvador,       \\& 
syria,           
rumania,         
bulgaria,        
iraq,            
balkans         
\vspace{1.5pt}\\
{\bf 1944 } &   
yugoslavia,      
salvador,        
iraq,            
guatemala,       
arabia,         \\& 
bulgaria,        
albania,         
rumania,         
syria,           
iran            
\vspace{1.5pt}\\
{\bf 1946 } &   
iraq,            
albania,         
arabia,          
salvador,        
guatemala,      \\& 
iran,            
bulgaria,        
afghanistan,     
rumania,         
syria           
\vspace{1.5pt}\\
{\bf 1948 } &   
iraq,            
albania,         
arabia,          
salvador,        
guatemala,      \\& 
iran,            
bulgaria,        
afghanistan,     
rumania,         
syria           
\vspace{1.5pt}\\
{\bf \blue{1950} } &   
\blue{iraq,            
arabia,          
albania,         
afghanistan,     
iran, }          \\& 
\blue{saudi,           
salvador,        
guatemala,       
ethiopia,        
cyprus}          
\vspace{1.5pt}\\
{\bf 1952 } &   
iraq,            
arabia,          
saudi,           
albania,         
afghanistan,    \\& 
iran,            
salvador,        
guatemala,       
cyprus,          
ethiopia        
\vspace{1.5pt}\\
{\bf 1954 } &   
iraq,            
albania,         
bulgaria,        
arabia,          
iran,            
salvador,       \\& 
afghanistan,     
rumania,         
syria,           
cyprus          
\vspace{1.5pt}\\
{\bf 1956 } &   
iraq,            
albania,         
iran,            
bulgaria,        
syria,          \\& 
rumania,         
afghanistan,     
salvador,        
arabia,          
guatemala       
\vspace{1.5pt}\\
 & \\ &
\vspace{1.5pt}
\end{tabular}
\end{minipage}\hfill
\begin{minipage}{.33\textwidth}
\begin{tabular}{ll}
\\
{\bf 1958 } &   
iran,            
syria,           
albania,         
afghanistan,     
iraq,           \\& 
bulgaria,        
arabia,          
rumania,         
cyprus,          
sultan          
\vspace{1.5pt}\\
{\bf 1960 } &   
iran,            
syria,           
albania,         
afghanistan,     
iraq,           \\& 
bulgaria,        
arabia,          
rumania,         
cyprus,          
sultan          
\vspace{1.5pt}\\
{\bf 1962 } &   
iran,            
iraq,            
syria,           
afghanistan,     
invasion,        \\&
invaded,         
indochina,       
egypt,           
cyprus,          
arabia          
\vspace{1.5pt}\\
{\bf 1964 } &   
iran,            
syria,           
iraq,            
afghanistan,     
invasion,        \\&
invaded,         
egypt,           
indochina,       
turkey,          
turkish         
\vspace{1.5pt}\\
{\bf 1966 } &   
iran,            
iraq,            
syria,           
invasion,        
afghanistan,     \\&
invaded,         
indochina,       
korea,           
egypt,           
aggressors      
\vspace{1.5pt}\\
{\bf 1968 } &   
iraq,            
iran,            
syria,           
invasion,        
invaded,         \\&
afghanistan,     
indochina,       
korea,           
aggressors,      
egypt           
\vspace{1.5pt}\\
{\bf 1970 } &   
iraq,            
iran,            
invasion,        
syria,           
invaded,         \\&
afghanistan,     
korea,           
indochina,       
aggressors,      
egypt           
\vspace{1.5pt}\\
{\bf 1972 } &   
iraq,            
invasion,        
iran,            
syria,           
invaded,         \\&
afghanistan,     
aggressors,      
korea,           
indochina,       
troops          
\vspace{1.5pt}\\
{\bf 1974 } &   
iraq,            
invasion,        
iran,            
korea,           
syria,           \\&
invaded,         
troops,          
aggressors,      
afghanistan,     
indochina       
\vspace{1.5pt}\\
{\bf 1976 } &   
iraq,            
iran,            
aggressors,      
aggression,      
syria,           \\&
invasion,        
troops,          
korea,           
invaded,         
indochina       
\vspace{1.5pt}\\
{\bf 1978 } &   
iraq,            
aggressors,      
troops,          
iran,            
invasion,        \\&
syria,           
korea,           
aggression,      
indochina,       
invaded         
\vspace{1.5pt}\\
{\bf \purple{1980} } &   
\purple{iraq,            
aggressors,      
iran,            
troops,          
invasion,}        \\&
\purple{syria,           
korea,           
aggression,      
indochina,       
invaded }        
\vspace{1.5pt}\\
{\bf 1982 } &   
iraq,            
aggressors,      
iran,            
troops,          
invasion,        \\&
syria,           
korea,           
aggression,      
indochina,       
invaded         
\vspace{1.5pt}\\
{\bf 1984 } &   
iraq,            
iran,            
aggressors,      
syria,           
invasion,        \\&
troops,          
korea,           
aggression,      
invaded,         
allies          
\vspace{1.5pt}\\
{\bf 1986 } &   
iraq,            
aggressors,      
syria,           
invasion,        
iran,            \\&
korea,           
troops,          
invaded,         
aggression,      
iraqi           
\vspace{1.5pt}\\
{\bf 1988 } &   
iraq,            
aggressors,      
invasion,        
iran,            
allies,          \\&
invaded,         
korea,           
iraqi,           
syria,           
aggression      
\vspace{1.5pt}\\
{\bf 1990 } &   
iraq,            
iraqi,           
iran,            
afghanistan,     
invaded,         \\&
aggressors,      
terror,          
allies,          
korea,           
invasion        
\vspace{1.5pt}\\
{\bf 1992 } &   
iraq,            
iran,            
terror,          
allies,          
iraqi,           \\&
korea,           
aggressors,      
afghanistan,     
syria,           
invaded         
\vspace{1.5pt}\\
{\bf 1994 } &   
iraq,            
iraqi,           
invaded,         
korea,           
allies,          \\&
aggressors,      
iran,            
exit,            
afghanistan,     
terror          
\vspace{1.5pt}\\
{\bf 1996 } &   
iraq,            
iran,            
iraqi,           
allies,          
afghanistan,     \\&
terror,          
invaded,         
korea,           
aggressors,      
syria           
\vspace{1.5pt}\\
{\bf 1998 } &   
iraq,            
terror,          
iran,            
iraqi,           
afghanistan,     \\&
occupation,      
allies,          
invaded,         
troops,          
invasion        
\vspace{1.5pt}\\
{\bf 2000 } &   
iraq,            
iraqi,           
afghanistan,     
terrorism,       
terror,          \\&
iraqis,          
iran,            
reconstruction,  
saddam,          
bosnia          
\vspace{1.5pt}\\
{\bf 2002 } &   
iraq,            
iraqi,           
afghanistan,     
terrorism,       
iran,            \\&
terror,          
iraqis,          
reconstruction,  
terrorist,       
terrorists      
\vspace{1.5pt}\\
{\bf 2004 } &   
iraq,            
iraqi,           
afghanistan,     
terror,          
terrorism,       \\&
iran,            
reconstruction,  
iraqis,          
terrorists,      
saddam          
\vspace{1.5pt}\\
{\bf 2006 } &   
iraq,            
iraqi,           
afghanistan,     
terror,          
terrorism,       \\&
iran,            
reconstruction,  
iraqis,          
terrorists,      
saddam          
\vspace{1.5pt}\\
{\bf \red{2008} } &   
\red{iraq,            
iraqi,           
afghanistan,     
terror,          
terrorism,}       \\&
\red{iran,            
reconstruction,  
iraqis,          
terrorists,      
saddam}         
\vspace{1.5pt}
\end{tabular}
\end{minipage}
\vspace{1.5pt}\\
\end{table*}
 \clearpage{}
\end{document}